\documentclass{article}
\usepackage[utf8]{inputenc}
\usepackage{multicol}
\usepackage[a4paper, total={6in, 8in}]{geometry}
\usepackage{comment}
\usepackage{graphicx}

\title{Applications of Generative Adversarial Models in Visual Search Reformulation}
\author{Kyle Xiao, Houdong Hu, Yan Wang}
\date{October 2019}

\begin{document}

\maketitle

\begin{multicols}{2}

\begin{abstract}

Query reformulation is the process by which a input search query is refined by the user to match documents outside the original top-n results. On average, roughly 50\% of text search queries involve some form of reformulation, and term suggestion tools are used 35\% of the time when offered to users. \cite{martihearst2009} As prevalent as text search queries are, however, such a feature has yet to be explored at scale for visual search. This is because reformulation for images presents a novel challenge to seamlessly transform visual features to match user intent within the context of a typical user session. In this paper, we present methods of semantically transforming visual queries, such as utilizing operations in the latent space of a generative adversarial model for the scenarios of fashion and product search.

\end{abstract}

\section{Introduction}
The usage of search engines has become the DeFacto method for navigation and information retrieval among internet consumers by allowing users to pivot and express intent using a search query. Particularly, visual search allows users to specify their intent through an image-based medium, which offers new dimensions of expressivity, nuance, and ease-of-use compared to traditional text-based queries. Technologies in visual search are still rapidly evolving to reflect emerging innovation and increasing consumer demand for low time to satisfaction, and is a popular and long standing research area \cite{DBLP:journals/corr/abs-1802-04914, Arandjelovic2012, Lazebnik, Smeulders2000, Yang2017, Zhai2017}.

A restriction of visual search, though, is that the ability for search suggestions and reformulations are limited. This is impractical for image queries since most image manipulation methods are difficult or infeasible in the context of a typical user session. In addition, existing filters and reformulation techniques fail to capture the full range of semantic expressivity a user may want to impart while preserving the original image intent. Our problem thus is how we combine a query image and semantic attribute into an image reformulation for a web browsing scenario.

Our solution to this problem leverages generative image models and conditioning the output of said models through novel optimization frameworks. That is, using models such as generative adversarial networks (GANs) \cite{1406.2661} to create high resolution and high fidelity synthetic images that match the original image query intent and a given semantic attribute. We target the fashion segment since this is currently a key investment opportunity for Bing shopping results and matches our problem paradigm closely. In addition, for the purposes of these experiments we limit the domain of images specifically to dresses to demonstrate the underlying concept.

The specific models and frameworks we use take advantage of the StyleGAN architecture \cite{1812.04948} and an augmentation to perceptual loss optimization \cite{1603.08155} in our encoder framework. In addition, we run analysis on the GAN latent space with respect to our semantic attributes, and use it to justify our novel gradient propagation framework for imparting semantic attributes. We also demonstrate the latency and modularity of our approach for use case in Bing search query pipelines.

We measure the performance of our reformulations using oracle normalized discounted cumulative gain (ONDCG) \cite{1304.6480}. This is done through human judges and ground truth reformulation collected from online catalogs like Amazon.com. We compare satisfaction of returned visual search results from Bing.com of the original image query, ground truth, and reformulation.

\section{Background and Related Work}

The corpus of work on generative image models is extensive. Deep generative deconvolutional networks (DGDNs) utilize stochastic unpooling to scale latent vectors, yielding a top-down image generation framework \cite{1512.07344}. In this paper, a Bayesian support vector machine is linked to the top-layer features,
yielding max-margin discrimination. Variational autoencoders build upon this, adding KL divergence loss to bias latent variables towards a Gaussian distribution \cite{1606.05908}. These methods, however, lack high enough fidelity and resolution for the purposes of the described reformulation paradigm.

GANs have been shown to be some of the most powerful generative image models available, and many GAN based architectures have been developed \cite{1709.06548, 1611.02200, 1611.07004}. CycleGAN provides a cycle-consistent loss framework for adapting images in different domains \cite{1703.10593}. CyCada extends this framework by adding both feature level, per pixel, and task loss \cite{1711.03213}. These frameworks, however, lack the level of granularity for attribute level adaptation in the image reformulation paradigm. AttGAN does manage to capture finer level attribute manipulation in the context of facial editing, but conditions the generator training on attribute classifications and makes few architectural changes to ensure unsupervised separation of high level attributes \cite{1711.10678}. The generator training in particular is impractical for shipping candidates in Bing since incremental and modular model training is an important feature, so a disjoint latent space optimization framework is preferable.

For the purpose of this paper, we leverage the state-of-the-art in generative image models in terms of distribution quality metrics. StyleGAN is a GAN architecture which uses a mapping network, AdaIN modules, stochastic variation, and truncation tricks \cite{1812.04948}. The mapping network allows a hierarchical representation of the latent vector to be incorporated into the generation step, which was shown to improve control of visual features and disentangle the latent space. The AdaIN modules transmits the encoded information w into the generated images at each resolution level while also controlling expression. For the purposes of this paper, the stochastic variation (Noise) is turned off since the original paper’s intent was to create variation in facial features, which is not relevant for image encoding and reformulation. Finally, the truncation on w eliminates outliers in the distribution of the latent space, thereby controlling areas of image generation that are poorly represented in the training data.

In addition, there has been extensive work on visual search on the Bing platform as well as numerous metrics proposed \cite{DBLP:journals/corr/abs-1802-04914}. Relevance-focused metrics have been shown to be the most representative measure for end user scenarios. For this reason we evaluate on ONDCG.

\section{Paradigm}

\noindent
\includegraphics[scale=0.17]{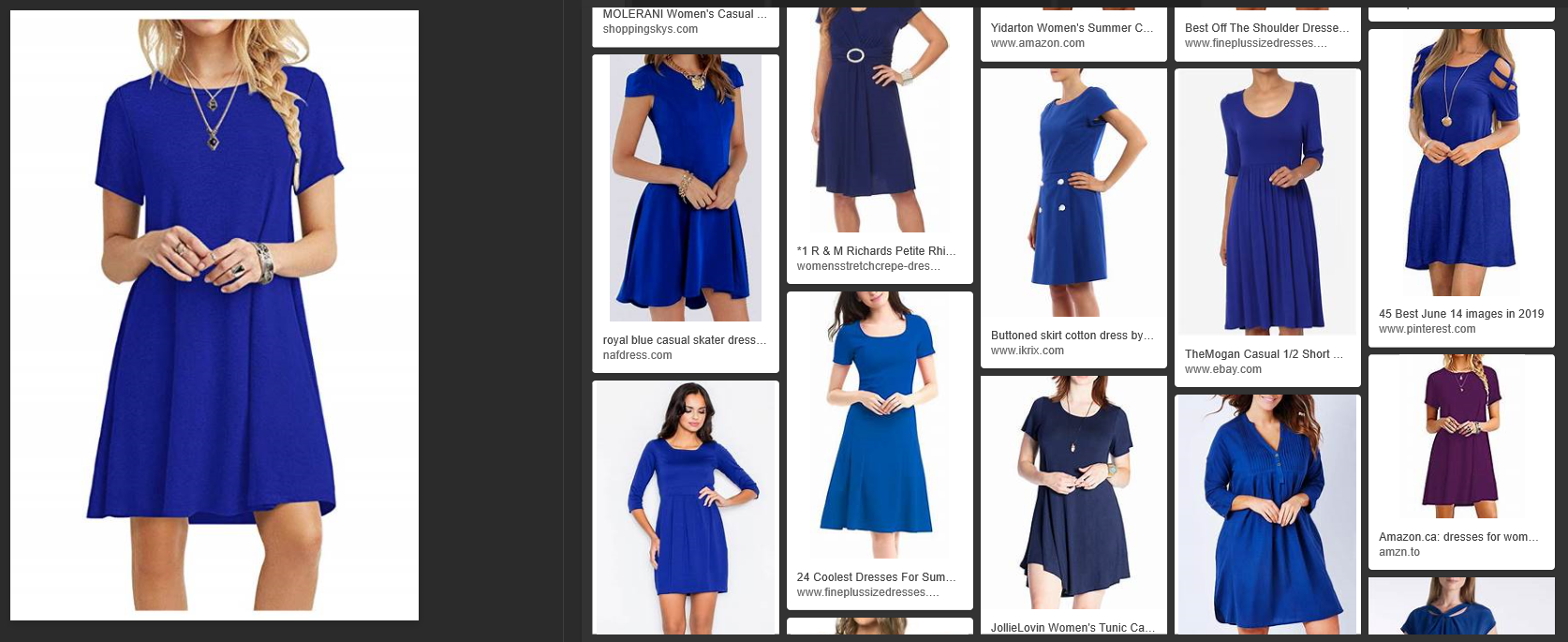}
\textit{Figure 1: Image search allows users to find dresses that are visually similar}\\
\indent
The current image search paradigm allows users to query visually similar images from a static image query known a priori by the user. This means that making edits on the fly is not possible. If a user wants the same dress with a floral pattern and in a different color, they must resort to external techniques like photoshop, likely needing to find a reference image with the desired attribute.\\

\noindent
\includegraphics[scale=0.45]{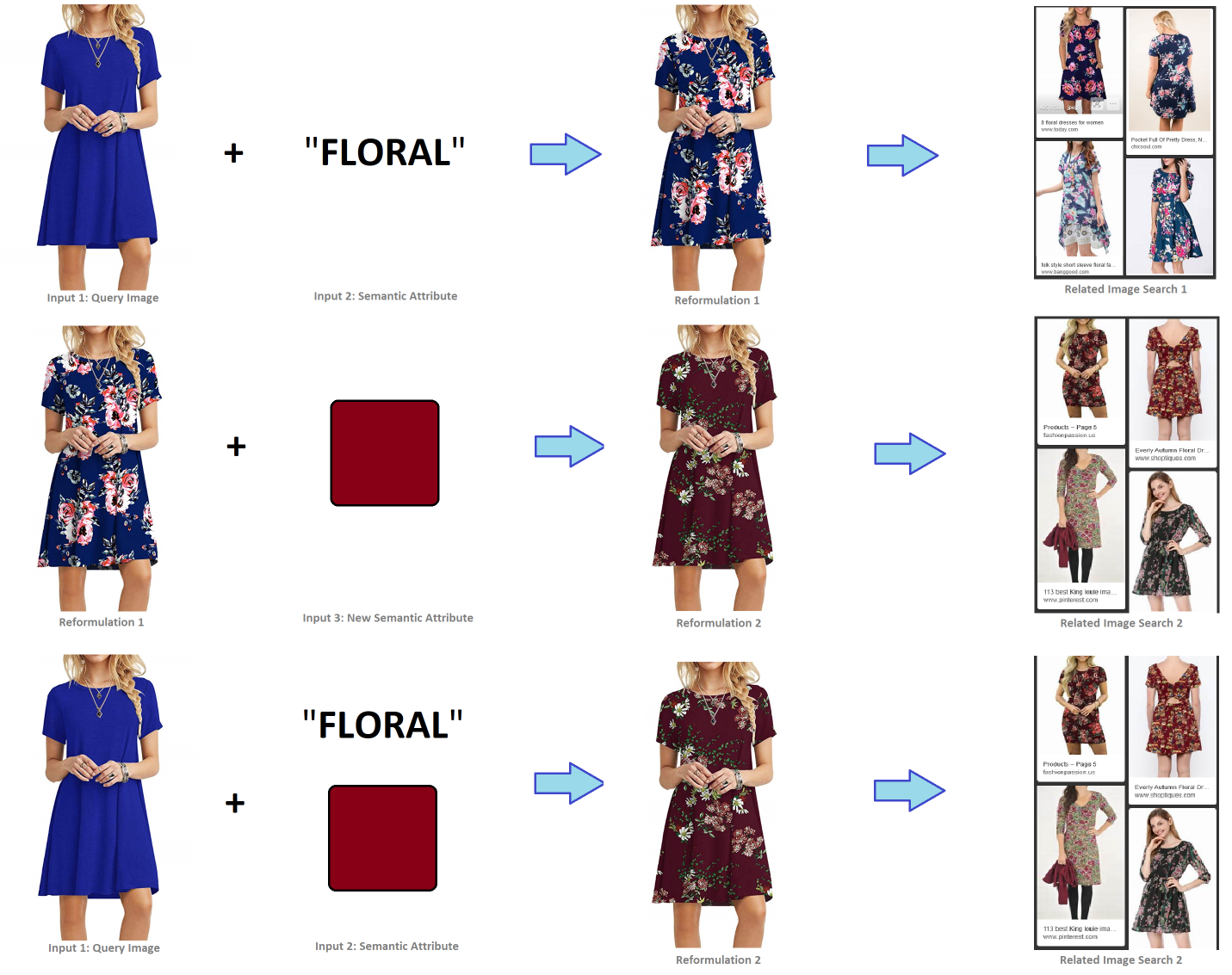}\\
\textit{Figure 2: The proposed paradigm for reformulation}\\

The proposed paradigm that this paper addresses takes an input image and a selected semantic attribute. In contrast to naive methods of image to image reformulation, the proposed method uses gradient propagation to translate latent variables, thus allowing for the paradigm shown where users can select a semantic attribute to base the transformation on without the need of a target image. In addition, multiple attributes can be mixed to enhance expressivity of reformulation. This is possible due to the recent development of Generative Adversarial Models (GANs) and the StyleGAN architecture, as well as the novel inclusion of gradient propagation transformations. Using these methods, it has been shown we can produce images with high enough quality and photo-realism to produce reasonable reformulations.

\section{Approach}

\noindent
\includegraphics[scale=0.5]{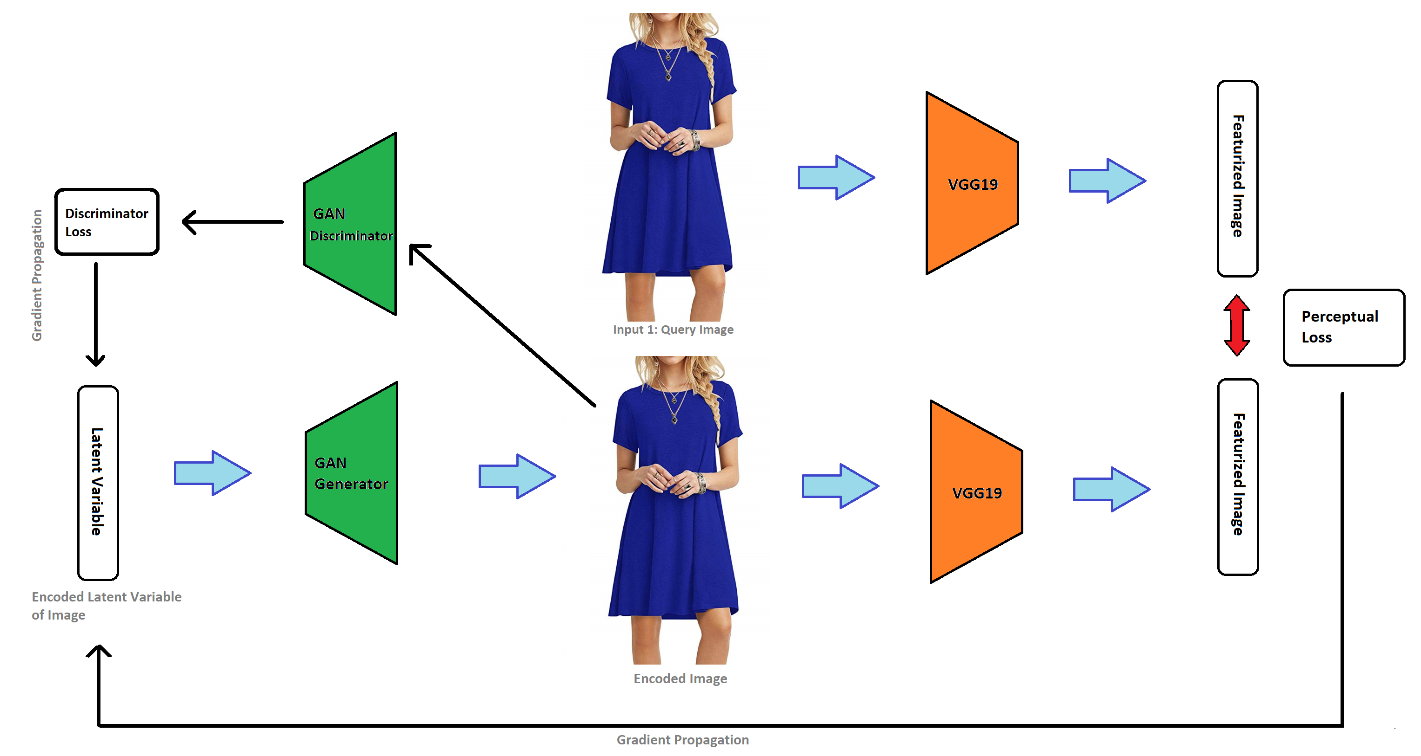}\\
\textit{Figure 3: Encoder framework}\\

\noindent
\includegraphics[scale=0.5]{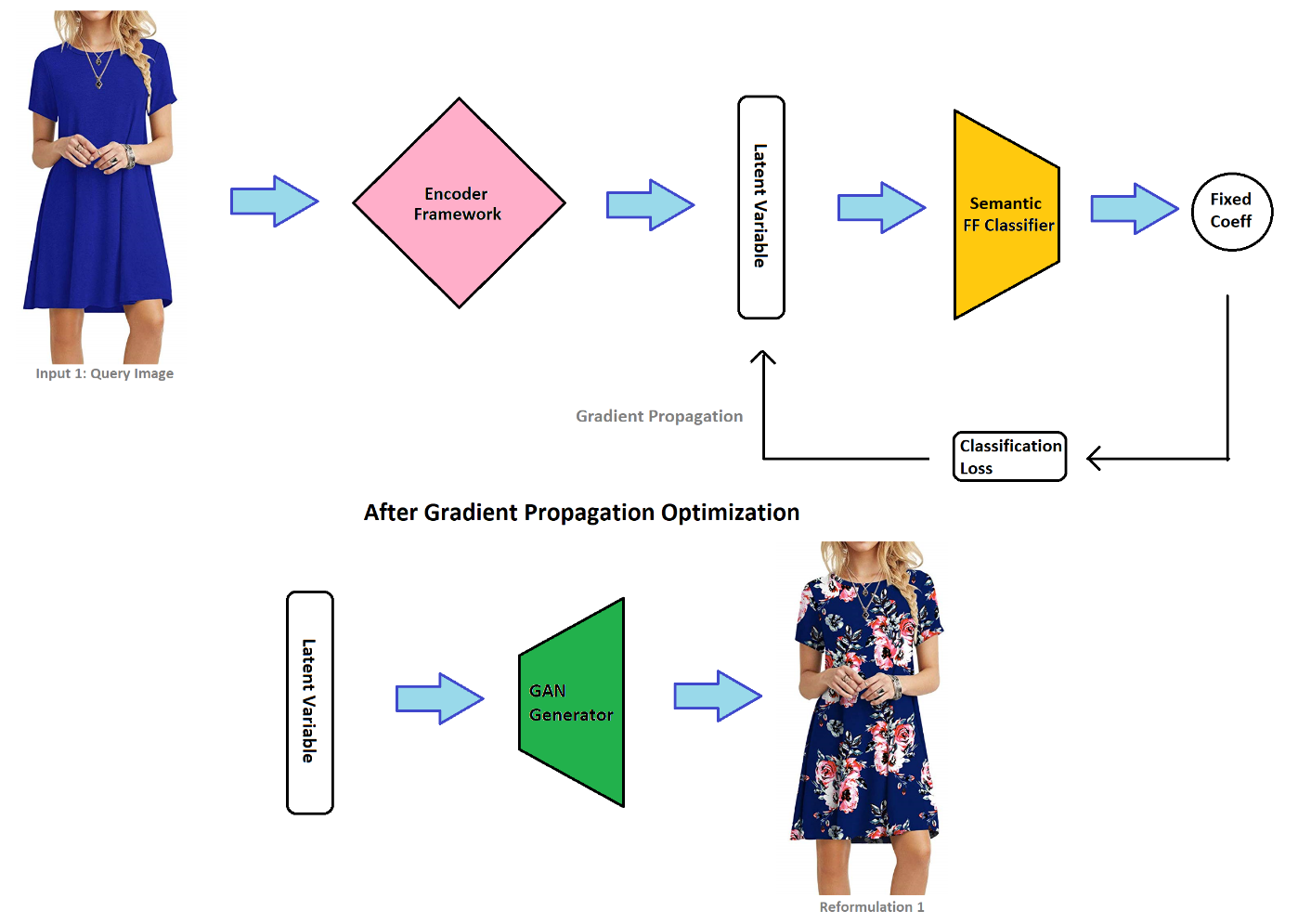}\\
\textit{Figure 4: Reformulation framework}\\

\indent
The framework of the described system is shown above in figures 3 \& 4.
The encoder framework takes as input an image and produces a GAN latent variable representation using perceptual loss optimization with discriminator loss. The reformulation framework propagates gradients from a semantic feedforward classifier trained on the desired attribute to the latent variable representation to move the encoded image closer to the desired attribute. The resulting latent variable has the semantic attribute imparted to it.

For the encoder framework, we first encode the images into GAN latent space by finding an optimal latent variable as follows:

\[{\mathop{\mathrm{min}}_{z} {\ \left\|V_i\left(x\right)-V_i\left(G(z)\right)\right\|}_2+\beta *D\left(G(z)\right)\ }\] 

Where z is the latent variable, x is the source image, G is the GAN generator, D is the GAN discriminator, Vi is the ith layer of VGG19, and $\beta$ is a tunable coefficient. In other words, we optimize for both perceptual loss and realism score from the discriminator. The actual minimization for this problem can be done with any optimization algorithm such as gradient descent or ADAM optimization. For latency reasons, z can also be estimated with a feedforward convolutional neural network trained on data optimized a priori:

\[z=CNN(x)\] \\

The reason for the encoding process is to transform the image into a representation that can be modified to add the semantic attribute. The latent variable is thus a 1xN vector which captures the semantic information of the image and can be modified to produce a semantically different image without having to do per pixel changes. 

\noindent
\includegraphics[scale=0.18]{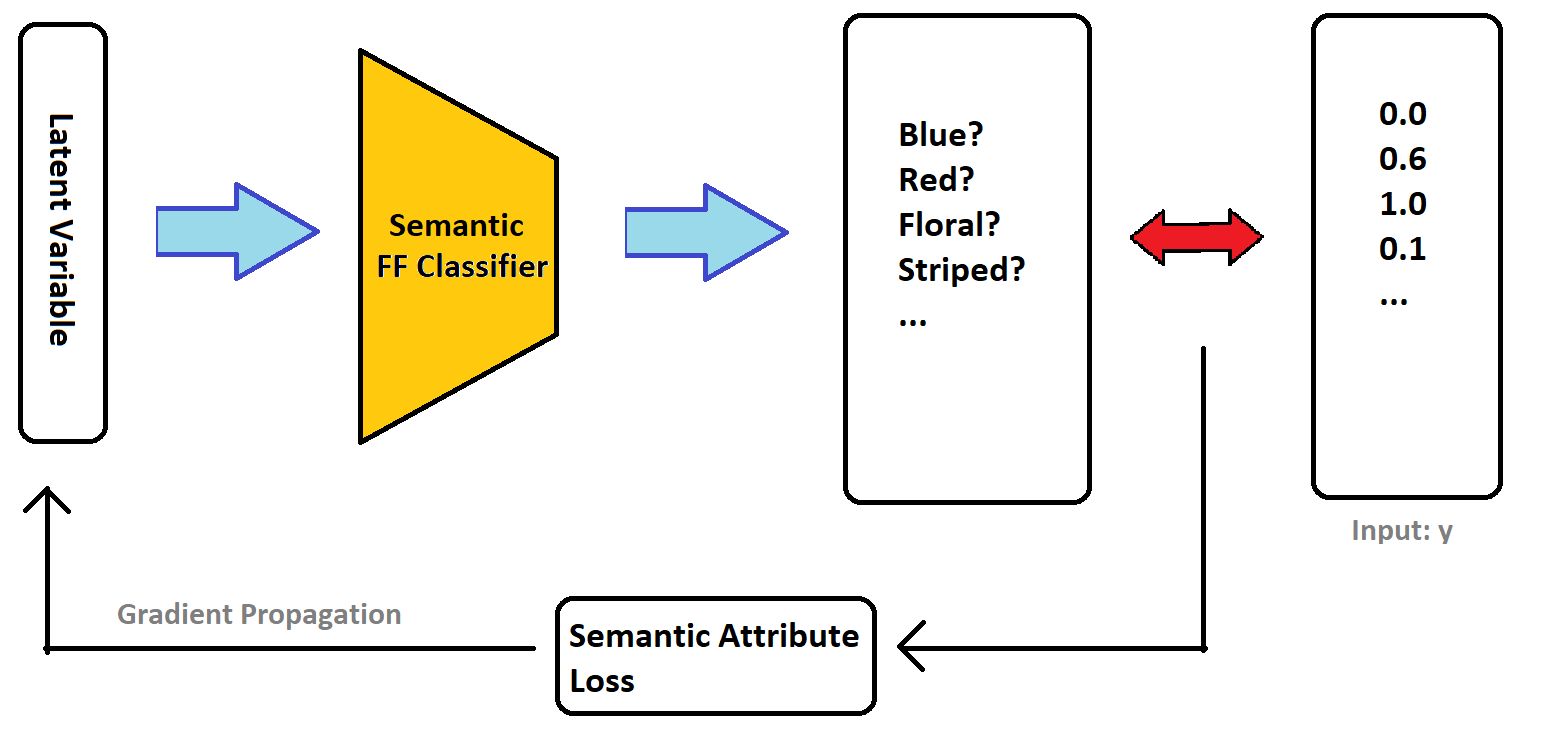}\\
\textit{Figure 5: The novel method for latent space transformation}\\

Once we have the encoding, we can make transformations on it using gradient propagation. The proposed approach utilizes using classifier gradient propagations. We first train a feedforward classifier for the desired attribute. Then, we then fix y to an arbitrary coefficient and propagate gradients to z.

\[z'=\mathop{\mathrm{min}}_{z}{\ \left\|y-FF(z)\right\|}_2,\ \ y\in (0,1)\]

\section{Measurement}

The proposed measurement utilizes oracle normalized discounted cumulative gain (ONDCG) \cite{1304.6480}. We first collect paired data from online catalogs like Amazon.com, where we have catalogs with known item differences. For instance, a single product listing may include the same product in different colors, prints, styles, etc. We pair products of the same listing but different attribute as the original image and ground truth for the reformulation of that attribute.

For a given product and attribute, we utilize human judges to compare returned visual search results for the original image, the reformulation, and the ground truth. This is done through discounted cumulative gain (DCG) analysis of the returned results \cite{1304.6480}, using a reference set composed of the original image queries. This way, we have a delta between the original image and ground truth image and can calculate ONDCG. We propose this as a measure of satisfaction of returned visual search results.

For the specific use case of color transformation and dress length on an online catalog of dress listings, our implmentation had an ONDCG score of 0.35 gain above the reference (n=50 listings, on average each m=5 attributes).

\section{Applications}

In addition to the application paradigms mentioned in Introduction, we propose the system can support other scenarios.

For instance, the system enables users to explore attributes by manually setting the value for y in the gradient propagation. Also, multiple semantic attributes are supported, as the semantic feedforward classifier can be used to classify using a one-hot encoding many different attributes. Fixing the output vector y could thus shape combinations of attribute features. In addition, the semantic attributes are general to support potentially pattern given sufficient data. For instance, scraping data for any arbitrary text query like “Taylor Swift dress” can result in a coherent classifier and gradient propagation.

\section{Conclusion}
In this paper, we proposed a novel method for visual search reformulation involving gradient propagation in GAN latent space. A user can control mixtures of semantic attributes so that he/she can look for a dress with “a floral pattern in a blue color.” These methods support a paradigm in which users can refine visual search queries using a range of given attributes without using a target image. We proposed the specific GAN framework, encoder structure, and latent space attribution methods as well as provided sample scenarios

\end{multicols}

\bibliographystyle{plain} 
\bibliography{sources} 

\end{document}